\documentclass[10pt,twocolumn,letterpaper]{article}

\usepackage{float}
\usepackage{iccv}
\usepackage{times}
\usepackage{epsfig}
\usepackage{graphicx}
\usepackage{amsmath}
\usepackage{amssymb}

\usepackage[breaklinks=true,bookmarks=false]{hyperref}

\iccvfinalcopy 

\def\iccvPaperID{****} 

\begin{document}

\title{Leveraging Model Interpretability and Stability to increase Model Robustness}

\author{Fei WU\\
CentraleSupelec, Valeo\\
{\tt\small fei.wu@supelec.fr}
\and
Thomas MICHEL\\
Valeo\\
{\tt\small thomas.michel@valeo.com}
\and
Alexandre BRIOT\\
Valeo\\
{\tt\small alexandre.briot@valeo.com}
}
\maketitle

\begin{abstract}
    State of the art Deep Neural Networks (DNN) can now achieve above human level accuracy on image classification tasks. However their outstanding performances come along with a complex inference mechanism making them arduously interpretable models. In order to understand the underlying prediction rules of DNNs, Dhamdhere \etal~\cite{conductance} propose an interpretability method to break down a DNN prediction score as sum of its hidden unit contributions, in the form of a metric called conductance. Analyzing conductances of DNN hidden units, we find out there is a difference in how wrong and correct predictions are inferred. We identify distinguishable patterns of hidden unit activations for wrong and correct predictions. We then use an error detector in the form of a binary classifier on top of the DNN to automatically discriminate wrong and correct predictions of the DNN based on their hidden unit activations. Detected wrong predictions are discarded, increasing the model robustness. A different approach to distinguish wrong and correct predictions of DNNs is proposed by Wang \etal~\cite{mutation} whose method is based on the premise that input samples leading a DNN into making wrong predictions are less stable to the DNN weight changes than correctly classified input samples. In our study, we compare both methods and find out by combining them that better detection of wrong predictions can be achieved.
\end{abstract}

\section{Introduction}

Deep Neural Networks (DNN) are known to be models that can achieve highly accurate performance on vision related tasks. Unlike conventional programming, rules followed by a DNN to process those tasks are not hard coded by hand but are instead learnt from the data information on which the DNN has been trained. However the end to end supervised learning method to train DNNs produces black box models and does not allow users to easily understand the logic of their inference. Debugging such complex systems in case of errors hence becomes a serious challenge. Understanding how DNNs make predictions has become crucial, all the more with the deployment of machine learning in decision critical applications and many researches have been conducted around the subject \cite{survey1, survey2, survey3}.
Methods resulting from these researches are successful in helping users understand why a given input is predicted as a specific output. Among those methods, there are so-called methods of attribution which break down the prediction score of a DNN as contribution of its previous units. For image classification tasks, these methods can allow users to visualize parts of the input image on which a DNN focused its attention to make a prediction \cite{SHAP, IG, deeplift}. They can also allow users to visualize which parts of a DNN hidden layers were most activation intensive in a given prediction \cite{deeplift, conductance, influence, fine_grained_text, decision_tree}.

Most of the interpretability methods in the literature, yet successful in making DNNs understandable, are limited to the stage of understanding DNN predictions and do not fully tackle the issue of debugging DNNs in addition to understanding them. In this work, we study a partial solution to debugging DNNs: instead of totally rectifying the prediction errors they made, we propose to detect whether the predictions of DNNs are wrong or not. By doing so, we reduce errors made by DNNs, increasing their robustness. The pipeline of this method is to first understand the underlying inference mechanism of a DNN, then test the inference logic of new inputs to check whether the DNN made a prediction using an appropriate reasoning. Among the previously cited interpretability methods, we choose to base our work on the one of Dhamdhere \etal \cite{conductance} as it stands out as the one to most accurately decompose the DNN prediction score as shown in their experiments. In the context of a DNN predicting an input sample, they define a value called conductance for each feature map of the DNN and break down the output pre-softmax score as the sum of all feature map conductances within a given layer. Using the decomposed expression of the pre-softmax score, they identify feature maps with the largest conductance value as the most contributive ones for this prediction. Then comparing the prediction decomposition of all images within the same class, they find out that there is a regular subset of feature maps with high conductance. They identify those feature maps as class specific feature maps, which are tasked to detect visual features representative of this class.

While experimenting with the conductance metric, we observe that it can be used to distinguish wrong and correct predictions of DNNs. Since Dhamdhere \etal show that predictions within a given class mostly rely on the activation of feature maps specific to this class (which also can be interpreted as detecting visual features specific to this class), we want to test the efficiency of their method by using the following experiment. \textbf{Given an image $x_A$ of class $A$ wrongly predicted into a class $B$ object, would feature maps specific to class $B$ react strongly to $x_A$ with a high conductance even though $x_A$ does not display visual features of class B?} The result is no, feature maps specific to class $B$ images do not react strongly to $x_A$ with high conductance as they do for most images of class $B$. On the one hand, this result supports the efficiency of Dhamdhere \etal's work and on the other hand, it enables us to define a method to discriminate wrong and correct predictions by breaking down their pre-softmax score as sum of feature map conductances.

We conduct experiments on two different random subsets of ImageNet dataset \cite{Imagenet} trained on an InceptionV3 DNN \cite{inception} and the full CIFAR10 dataset \cite{Cifar10} trained on a ResNet20 DNN \cite{resnet}. These datasets are split into training, validation and test sets. Regarding predictions as a given class, we pay attention to the conductance of all feature maps and not only the ones with the highest conductance. We observe that there is a different pattern of conductance value between correct and wrong predictions. We decide from there to learn a binary classifier (acting as an error detector) on top of the DNNs to automatically classify their predictions as wrong or correct. The binary classifier is trained on conductances computed from training sets and evaluated on conductances computed from validation and test sets. We measure the performance of the binary classifier with the Area Under the Receiver Operating Characteristic curve (AUROC). On one of the two ImageNet subsets for example, the binary classifier achieves an AUROC of 0.941 on the validation set (AUROC of 1 is a classifier without errors). In essence we propose to add a binary classifier beyond the original DNN. After the DNN made a prediction on an input sample, we compute the conductance pattern associated to this prediction. The binary classifier then process this conductance pattern to decide whether it looks like a wrong prediction pattern for the predicted class or not. Our code is available on \href{https://github.com/feiwu77777/Leveraging-Model-Interpretability-and-Stability-to-increase-Model-Robustness}{github}.

\section{Related works}

We present a method to increase the robustness of a Deep Neural Network (DNN) by discriminating its wrong and correct predictions. However there already exist methods in the literature to discriminate wrong and correct predictions of a DNN \cite{mutation, input_pert, density_estimate}. These methods nevertheless differ from our approach as they do not harness the interpretability of DNNs. Among previous works the one of Wang \etal \cite{mutation} draws our attention as they show the superiority of their method compared to the others in their experiments. In their work, they focus on distinguishing adversarial samples from normal samples but they show that they can also use their method to distinguish wrong from correct predictions in a more general way. By considering DNNs as black boxes, they manage to find a distinction between wrong and correct predictions using a measure of stability. They define the \textbf{Label Change Rate} (LCR) as the fraction of prediction changes among a set of slightly modified DNNs w.r.t. an original one. They observe that input samples leading the original DNN into wrong predictions often have their prediction results differ from the original ones when inferred from the modified DNNs (high LCR). By contrast, for correctly predicted input samples, the predictions made by modified DNNs are often consistent with the results of the original one (low LCR). 

They adopt a statistical hypothesis testing approach \cite{stat_testing1, stat_testing2} and show through their experiments that LCR can accurately detect wrong predictions made by LeNet \cite{LeNet} and GoogLeNet \cite{googlenet} DNNs trained over MNIST and CIFAR10 dataset respectively. Differently from them, we adopt a simplified threshold approach for the LCR, so that this metric can be comparable with the metric based on the conductance value.
While comparing both metrics, we observe that conductance is generally faster and achieves better results than LCR with the simplified threshold approach. Furthermore, we also find out that the combination of LCR with conductance can achieve better results on classifying the predictions of the DNN as wrong or correct. Because the LCR and conductance methods deal with the same (wrong and correct prediction distinction) problematic in two inherently different ways (one considers the DNN as a black box while the other considers the DNN as a white box), we assume that the combination of both methods to detect wrong predictions can be more accurate than if they were used separately. We test our hypothesis on the previous three datasets and successfully increase the performance of the binary classifier as measured by the AUROC.

\section{Neuron Conductance}

We now present in greater detail the previously introduced conductance method for predicting whether the output of a Convolutional Neural Network (CNN) is correct or not. In contrast to LCR, this method relies on Neuron Conductance, which is a white box approach, requiring to keep track of each internal activation of the CNN. Convolutional layer outputs are 3D tensors of 2D feature maps and each item of a feature map is defined as neuron thereafter.

\subsection{Backgrounds}

Given a CNN making a prediction on an input image, Dhamdhere \etal present a method to break down the pre-softmax score of the predicted class as the contribution of neurons within a given convolutional layer. They build their research on the work \cite{IG} of Sundararajan \etal who define a method to break down the pre-softmax score as the contribution of the input image pixels:

\footnotesize
\begin{equation} \label{equation_IG}
\sum_{i} (x_i - x'_i)\int_{\alpha = 0}^{1} \frac{\partial f_{pred}(\gamma(\alpha))}{\partial \gamma_i(\alpha)} d\alpha = f_{pred}(x) - f_{pred}(x')
\end{equation}
\normalsize

where:
\begin{itemize}
  \item $x$ is the input image and $x_i$ is its $i^{th}$ pixel value,
  \item $x'$ is a reference input (all 0 valued image for example),
  \item $\gamma(\alpha) = x'+\alpha(x-x')$ is an intermediate image between the reference image and the input image (and $\gamma_i(\alpha) = x_i'+\alpha(x_i-x_i'))$,
  \item $f_{pred}$ is the forward pass function of a DNN which outputs the pre-softmax activation value at the predicted $pred$ class.
\end{itemize}

The conductance of a neuron $y$ w.r.t a predicted class $pred$ is then defined as:

\scriptsize
\begin{equation} \label{equation_NC}
    Cond^y_{pred}(x) = \sum_{i} (x_i - x'_i)\int_{\alpha = 0}^{1} \frac{\partial f_{pred}(\gamma(\alpha))}{\partial f_y(\gamma(\alpha))}\times \frac{\partial f_y(\gamma(\alpha))}{\partial \gamma_i(\alpha)}d \alpha
\end{equation}
\normalsize

where $f_y$ is a forward pass function which outputs the activation value of neuron $y$ and for any convolutional layer $L$, the following relation is verified:

\begin{equation} \label{equation_cond&presoftmax}
    \sum_{y \in L} Cond^y_{pred}(x) = f_{pred}(x) - f_{pred}(x') 
\end{equation}

Dhamdhere \etal then \textbf{define the conductance of a feature map as the sum of its neuron conductances}. The previous pre-softmax score can be rewritten as a contribution of feature map conductances, and feature maps with the highest conductance are regarded as the most contributive feature maps to this prediction. Other metrics (activation, activation$\times$gradient, internal influence \cite{influence}) also exist for capturing most relevant feature maps of a prediction, but Dhamdhere \etal show the superiority of their metric through experiments described in detail in their paper \cite{conductance}. They nevertheless stated an issue of scalability with their conductance formula as it loops through every pixel of the input image hence being computationally expensive. Schrikumar \etal solve this issue by suggesting a faster computation of the conductance. The calculation details are reported in their paper \cite{faster_cond} and their research results in the following formula:

\small
\begin{equation} \label{equation_fastcond}
    Cond^y_{pred}(x) = \int_{\alpha = 0}^{1} \frac{\partial f_{pred}(\gamma(\alpha))}{\partial f_y(\gamma(\alpha))} \times\frac{\partial f_y(\gamma(\alpha))}{\partial \alpha}d \alpha
\end{equation}
\normalsize

In order to compute the calculation of the conductance, a $n$ ($n = 10$ in the later experiments) steps Riemann approximation of the integral is used, resulting in the following equation:

\small
\begin{equation} \label{equation_fastcond_approx}
    Cond^y_{pred}(x) = \sum_{k=1}^{n}\frac{\partial f_{pred}(x^{(k)})}{\partial f_y(x^{(k)})}\times(f_y(x^{(k)}) - f_y(x^{(k-1)}))
\end{equation}
\normalsize

with $x^{(k)} = \gamma(\frac{k}{n}) = x' + \frac{k}{n}(x-x')$.

\subsection{Observation}

\begin{figure}[t]
\begin{center}
\includegraphics[width=1.0\linewidth]{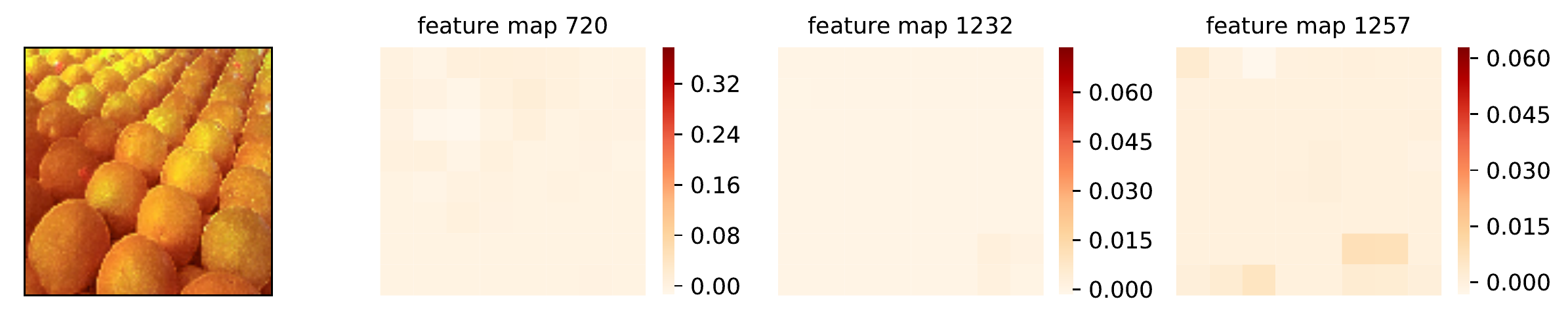}
\includegraphics[width=1.0\linewidth]{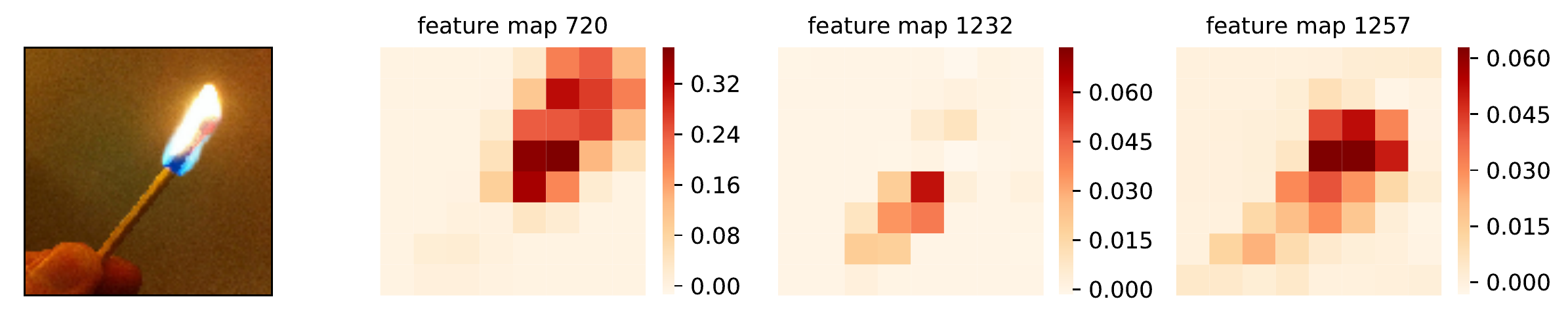}
\end{center}
   \caption{Orange and matchstick images both predicted as \textit{matchstick} with respectively 0.991 and 0.999 softmax scores displaying clear neuron conductance value difference on some feature maps of $FM^{high}_{(matchstick, mixed8)}$.}
\label{figure_fmcond}
\end{figure}

When comparing the pre-softmax score decomposition of correctly predicted images within a given class and for a given convolutional layer, Dhamdhere \etal find out that there are regularly the same subset of feature maps with higher conductance than the rest. They then label those feature maps, feature maps most specific to this class and \textbf{each one of them is specialized in detecting visual features related to this class of images}. Given a class $c$ and a convolutional layer $L$ of a CNN, $FM^{high}_{(c,L)}$ is the subset of high conductance valued feature maps from layer $L$ for this class. If images from other classes were incorrectly predicted as $c$ images, empirical observation shows that these images often have lower conductance value on feature maps of $FM^{high}_{(c,L)}$ than $c$ images. We show in Tables \ref{table_cond_pos} and \ref{table_cond_neg} the difference of conductance value observed in the layer \textit{mixed8} of the InceptionV3 CNN through four classes from the Imagenet dataset: \textit{matchstick}, \textit{orange}, \textit{stingray} and \textit{bison}. In Figure \ref{figure_fmcond}, we show for a specific example of an \textit{orange} image predicted as \textit{matchstick} and a \textit{matchstick} image predicted as a matchstick that checking on conductance value of feature maps from $FM^{high}_{(c,\textit{mixed8})}$ easily allows us to distinguish which one is the wrong prediction and which one is the correct prediction.

\begin{table}[t]
\centering

\scalebox{0.66}{
\begin{tabular}{|c|c|c|c|}
\hline
\textbf{matchstick} & \textbf{orange} & \textbf{stingray} & \textbf{bison} \\
\hline
\begin{tabular}{c|c}
     \textbf{FM} & \textbf{mean}\\
     \hline
     720  & 1.62 \\
     1232 & 8.32e-1 \\
     1257 & 7.23e-1\\
\end{tabular}&
\begin{tabular}{c|c}
    \textbf{FM} & \textbf{mean}\\
    \hline
    1204&9.41e-1\\
    615&9.10e-1\\
    521&8.38e-1\\
\end{tabular}&
\begin{tabular}{c|c}
    \textbf{FM} & \textbf{mean}\\
    \hline
    1261&1.03\\
    577&5.53e-1\\
    115&5.01e-1\\
\end{tabular}&
\begin{tabular}{c|c}
    \textbf{FM} & \textbf{mean}\\ 
    \hline
    1007&1.89\\
    574&6.23e-1\\
    1104&5.46e-1\\
\end{tabular}\\
\hline
\end{tabular}}
\caption{Mean conductance of feature maps from $FM^{high}_{(c,\textit{mixed8})}$ obtained by averaging conductance of all images correctly predicted as the corresponding class ordered decreasingly.}
\label{table_cond_pos}

\scalebox{0.66}{
\begin{tabular}{|c|c|c|c|}
\hline
\textbf{matchstick} & \textbf{orange} & \textbf{stingray} & \textbf{bison} \\
\hline
\begin{tabular}{c|c}
     \textbf{FM} & \textbf{mean}\\
     \hline
     720  & 2.47e-1\\
     1232 & 1.82e-1 \\
     1257 & 6.85e-1\\
\end{tabular}&
\begin{tabular}{c|c}
    \textbf{FM} & \textbf{mean}\\ 
    \hline
    1204&3.67e-1\\
    615&4.60e-1\\
    521&4.00e-1\\
\end{tabular}&
\begin{tabular}{c|c}
    \textbf{FM} & \textbf{mean}\\ 
    \hline
    1261&7.04e-1\\
    577&5.08e-1\\
    115&4.81e-1\\
\end{tabular}&
\begin{tabular}{c|c}
    \textbf{FM} & \textbf{mean}\\
    \hline
    1007&6.53e-1\\
    574&1.86e-1\\
    1104&6.37e-1\\
\end{tabular}\\
\hline
\end{tabular}}
\caption{Mean conductance of feature maps from $FM^{high}_{(c,\textit{mixed8})}$ obtained by averaging conductance of all images incorrectly predicted as the corresponding class.}
\label{table_cond_neg}
\end{table}

\begin{figure}[t]
\begin{center}
\includegraphics[width=1.0\linewidth]{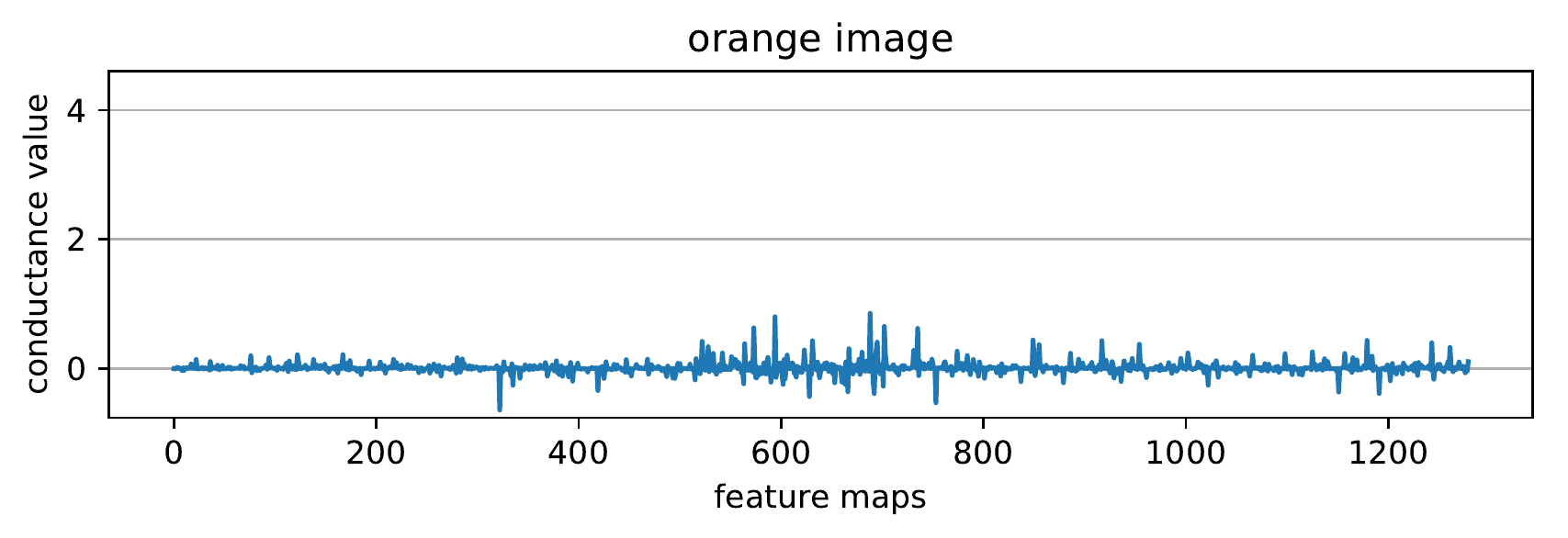}
\includegraphics[width=1.0\linewidth]{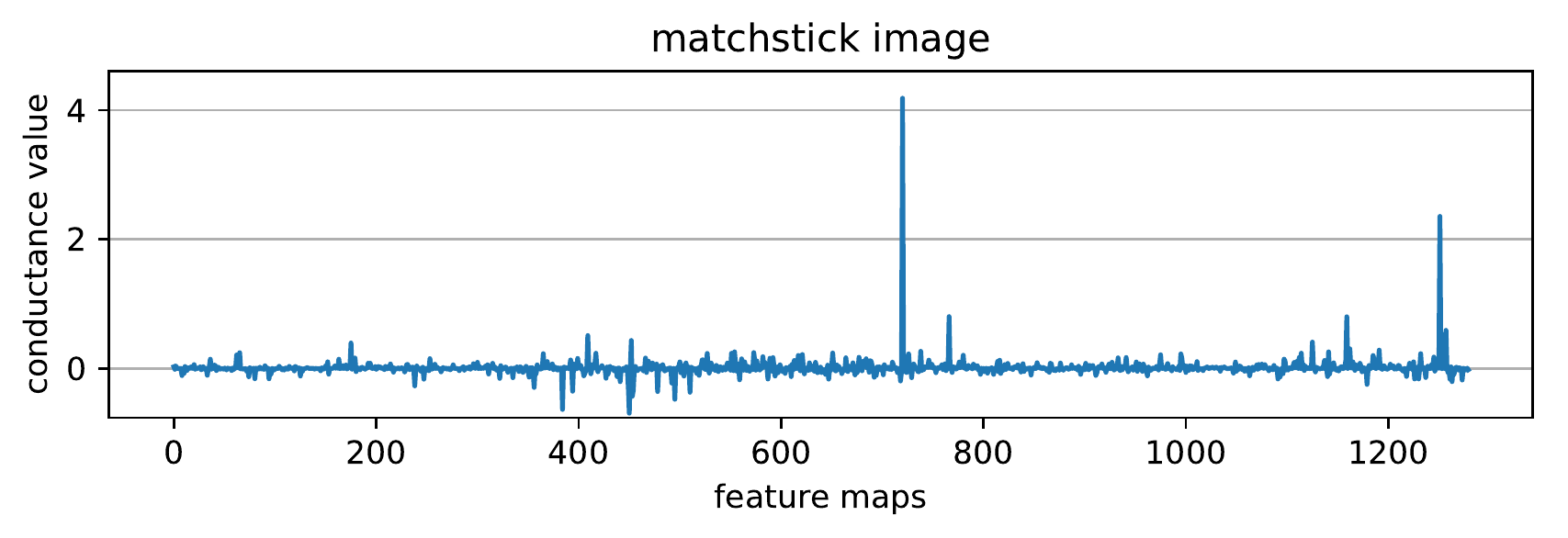}
\end{center}
   \caption{Feature map conductance values of an orange and a matchstick images (both predicted as \textit{matchstick} with respectively 11.7 and 15.8 pre-softmax score) on all feature maps of layer \textit{mixed8} from the InceptionV3 network.}
\label{figure_allcond}
\end{figure}

Using \textit{matchstick} specific highest conductance valued feature maps to distinguish wrong and correct matchstick predictions can be interpreted as checking if the CNN predicted a matchstick by emphasizing matchstick features. However looking at feature maps other than those with high conductance can be worthwhile too. For example given a feature map that is usually low conductance valued, for it to have high conductance when predicting a new image can be a hint of error. While experimenting with the method of Dhamdhere \etal we further observe for correct predictions that their pre-softmax decomposition often attribute comparable conductance values on every feature maps. Then when analyzing the conductance pattern of wrong predictions pre-softmax decomposition, we observe that there is also a regularity between pattern of wrong predictions. However their conductance pattern is nevertheless different from the pattern of correct predictions. An example of that pattern difference on feature maps of the InceptionV3 layer \textit{mixed8} for previous images of \textit{orange} and \textit{matchstick} both predicted as \textit{matchstick} is presented in Figure \ref{figure_allcond}. We observe that the matchstick image prediction (correct) decomposition is more focused on a few feature maps while the orange image prediction (wrong) decomposition is more indecisive and have a more averaged distribution of conductance value across all feature maps. For a correct and an incorrect prediction as a given class, their pre-softmax score $F$ can be of close value, making them hardly distinguishable which is the case of the matchstick ($F = 15.8$) and orange ($F = 11.7$) images. However breaking down $F$ as the sum of feature map conductances can give us a clearer insight on which one is the correct prediction and which one is the wrong prediction.

\section{Experiments}

The objective of these experiments is to show by adding a binary classifier (error detector) beyond the classic image classification pipeline, that we can detect and discard some of the wrong predictions made by a Convolutional Neural Network (CNN). We conduct experiments on three datasets. The first one is the CIFAR10 dataset used to train a ResNet20 CNN, the second one is a subset of 50 classes randomly selected from the ImageNet dataset used to train an InceptionV3 CNN and the last one is another random subset of ImageNet with this time 100 classes also used to train an InceptionV3. We split each one of previous datasets into three parts: training, validation and test sets. \textbf{Conductance computed from images of the training, validation and test sets are used as training, validation and test sets for the binary classifier}. For CIFAR10, the 50.000 images of the original training set are kept as such and the test set of 10.000 images is evenly split into validation and test sets. For the 50-classes ImageNet, 500 and 50 images per class are randomly extracted from the original training set of 1300 images per class to form the new training and validation sets, the original test set of 50 images per class remains unchanged. For the 100-classes ImageNet, 1150 and 150 images are extracted from the original training set to form the new training and validation sets and the original test set remains unchanged. We train the ResNet20 from scratch with the CIFAR10 dataset, after approximately 60 epochs, it achieves 86.2\%, 82.9\% and 81.9\% accuracy on the training, validation and test sets respectively. An InceptionV3 CNN is fine-tuned with the 50-classes ImageNet dataset, after 2 epochs, it achieves 97.8\%, 98.1\% and 93.0\% on the training, validation and test sets respectively. Another InceptionV3 CNN is fine-tuned with the 100-classes ImageNet dataset, after 2 epochs, it achieves 93.5\%, 90.9\% and 88.9\% on the training, validation and test sets respectively.

\begin{figure}[t]
\begin{center}
\includegraphics[width=0.9\linewidth]{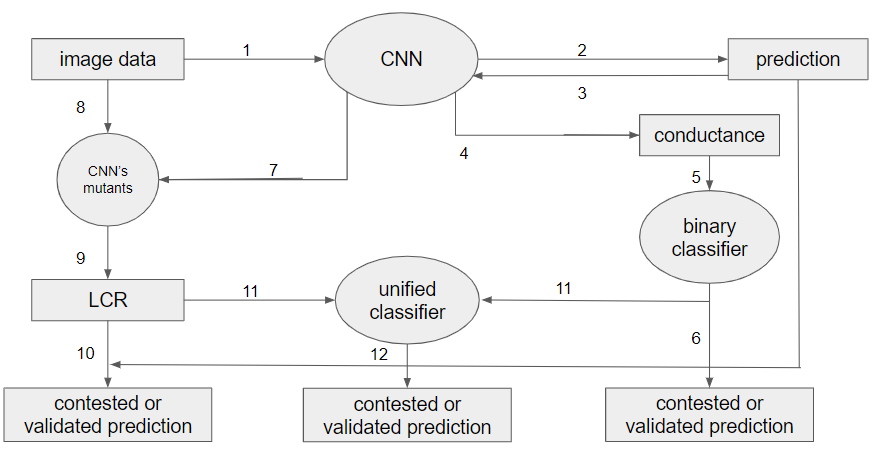}
\end{center}
\caption{Steps to compute conductance and LCR from the image dataset and use them to contest or validate predictions of the CNN.}
\label{figure_system}
\end{figure}

\subsection{Conductance data}
\subsubsection{From image to conductance}

Conductance of a CNN feature map is a metric that can be computed if we know the input image and the associated prediction made by the CNN for this image (steps 1 and 2 of Figure \ref{figure_system}). As there are several convolutional layers in the InceptionV3 and ResNet20 CNNs, we restrict the conductance computation to only the last convolutional layer feature maps because experiments show that the binary classifier would perform as much if not better by using only conductances of this layer. To form the conductance training set of the binary classifier, samples from the image training set are used to calculate neuron conductances according to equation \ref{equation_fastcond_approx}. Resulting neuron conductances are summed over space into feature map conductances (steps 3 and 4) and used as training set for the binary classifier. In the same manner, conductance computed from samples of the image validation set and image test set become the conductance validation set and conductance test set of the binary classifier. For the InceptionV3 CNN, since the last convolutional layer \textit{mixed10} has 2048 feature maps, an image is represented by  a 2048 long vector with each value of the vector corresponding to each feature map's conductance. Ground truth labels used for the binary classifier (error detector) are either 1 for wrong predictions or 0 for correct ones.

\subsubsection{Binary classifier training and results}

After computing conductance characteristics from the images samples and their associated predictions, we need to analyze them with a classifier in order to identify conductance patterns linked to wrong or correct predictions. We use a fully connected neural network of two layers with 200 hidden units per layer as the binary classifier. In the original pipeline of image classification, there are a lot more correct predictions than wrong predictions resulting in highly unbalanced learning samples for the binary classifier. In order to solve this issue, we use a ROC curve after training to choose a threshold which is adapted in separating both classes efficiently. During training phase, we also use the AUROC to monitor the performance of the binary classifier instead of the accuracy metric. After training the binary classifier on each conductance training set of the three datasets (step 5), we evaluate the binary classifier’s performance on the respective conductance validation set. The ROC curves obtained from results of the conductance validation sets are displayed in Figure \ref{figure_allROCs}. ROC curve is a commonly used tool to evaluate the performance of binary classifiers. With a threshold varying from 1 to 0, the recall of class 0 (percentage of the CNN correct predictions classified as correct by the binary classifier) and the recall of class 1 (percentage of the CNN wrong predictions classified as wrong) are calculated and define a point on the ROC curve. Overall, the more the area under the ROC curve is close to 1, the more accurate the binary classifier is in separating both classes. Depending on the context in which the binary classifier is used, different thresholds can be chosen so that the binary classifier can privilege performances on one of the two classes or not. For our following experiments, \textbf{we choose a threshold that enables the binary classifier to have close recall scores for both classes on the validation set}. By doing so, the percentage of un-discarded correct predictions (among total correct predictions) will be close to the percentage of discarded wrong predictions (among total wrong predictions). Thus for each of the three validation sets, we use the previous method to select a threshold (marked by a circle in Figure \ref{figure_allROCs}) and display the binary classifier’s results on their respective validation and test sets (step 6) in Tables \ref{table_cifar10}, \ref{table_50imagenet} and \ref{table_100imagenet}.

\subsection{Label Change Rate}
\subsubsection{Backgrounds}

We now present in this section how the Label Change Rate (LCR) characteristic of input images is computed and used to discriminate wrong and correct predictions. Label Change Rate is originally designed by Wang \etal to detect adversarial samples. However they show that their method can also be used to detect wrong predictions in a more general way. Given an original CNN $f$, they modify a random small subset of its weights (with one weight defined as a convolutional kernel within a convolutional layer) to obtain a CNN mutant. When generating mutants, only those that achieve at least 90\% of the original CNN's accuracy on the validation set are kept. After creating a set of mutants $F$, given an input image $x$, for each mutant $f^{(i)}\in F$ they make a prediction $f^{(i)}(x)$ on $x$ and calculate the LCR of this image with the following equation:
\begin{equation} \label{equation_LCR}
    LCR(x) = \frac{|f^{(i)} \in F, \quad f^{(i)}(x) \neq f(x)|}{|F|}
\end{equation}
where $|S|$ denotes the number of elements in a set $S$ and $f(x)$ is the prediction of the original CNN on $x$. They show that images leading the original CNN to make wrong predictions have a much higher LCR than images which do not. Wang \etal adopt a statistical hypothesis testing approach w.r.t. a targeted level of confidence to decide whether a given input $x$'s prediction is wrong or not. Differently from them, we adopt a simplified threshold based approach. We choose to use such an approach for the LCR metric so that it can be compared with the conductance metric. We first generate a fixed number of mutants (step 7 of Figure \ref{figure_system}), we then compute the LCR of all samples from the training, validation and test sets using equation \ref{equation_LCR} (steps 8 and 9). Finally, we choose a threshold with the ROC curve created from the validation set LCR values to make decisions on the test set LCR values.

\subsubsection{Hyper-parameter choices and results}

\begin{figure}[t]
\begin{center}
\includegraphics[width=0.6\linewidth]{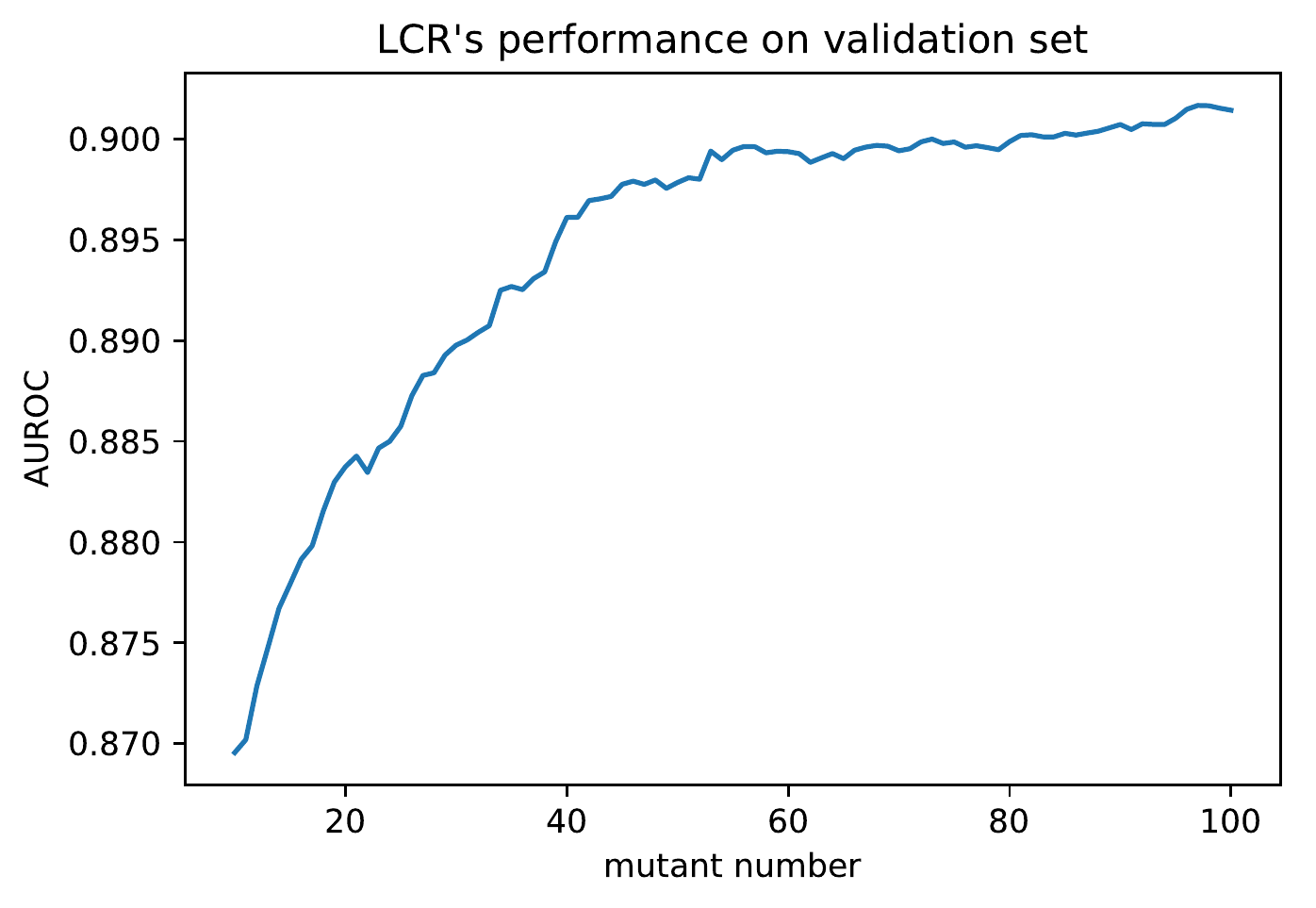}
\end{center}
\caption{LCR performance on the 100-classes ImageNet validation set depending on the number of mutants.}
\label{figure_auroc_mn}
\end{figure}

\begin{table}[t]
\centering
\scalebox{0.60}{
\begin{tabular}{|c|c|c|c|c|}
\hline
Operator& Mutation Rate& Correct Predictions& Wrong predictions&LCR difference\\
\hline
NAI&
\begin{tabular}{c}
     0.15\%  \\
     0.2\% \\
     0.25\%\\
\end{tabular}&
\begin{tabular}{c}
     5.37e-2 $\pm$ 1.20e-1  \\
     7.11e-2 $\pm$ 1.40e-1  \\
     8.00e-2 $\pm$ 1.50e-1 \\
\end{tabular}&
\begin{tabular}{c}
     3.63e-1 $\pm$ 2.24e-1  \\
     4.15e-1 $\pm$ 2.30e-1  \\
     4.32e-1 $\pm$ 2.36e-1  \\
\end{tabular}&
\begin{tabular}{c}
     -3.67e-2  \\
     -2.61e-2 \\
     -3.40e-2  \\
\end{tabular}
\\
\hline
GF&
\begin{tabular}{c}
     0.15\%  \\
     0.2\% \\
     0.25\%
\end{tabular}&
\begin{tabular}{c}
     1.55e-2 $\pm$ 5.89e-2 \\
     1.93e-2 $\pm$ 6.69e-2 \\
     2.54e-2 $\pm$ 7.63e-2 \\
\end{tabular}&
\begin{tabular}{c}
     1.85e-1 $\pm$ 1.83e-1  \\
     2.25e-1 $\pm$ 2.01e-1 \\
     2.45e-1 $\pm$ 2.06e-1 \\
\end{tabular}&
\begin{tabular}{c}
     -7.24e-2  \\
     -6.22e-2  \\
     -6.27e-2  \\
\end{tabular}
\\
\hline
\end{tabular}}
\caption{Average and standard deviation of the 100-classes ImageNet LCR for several operators and mutation rates. The LCR difference metric is calculated by subtracting the lower mean of wrong predictions by the higher mean of correct predictions.}
\label{table_LCR}
\end{table}

There are three hyper-parameters to take into account: the type of operation used to create a mutant, the percentage of weights to modify in the original CNN and the number of mutants to generate. To modify the original CNN’s weights, Wang \etal use four operators: \textit{Gaussian Fuzzing}, \textit{Neuron Activation Inverse}, \textit{Weight Shuffling} and \textit{Neuron Switch} among which \textit{Gaussian Fuzzing} and \textit{Neuron Activation Inverse} are recommended by the authors as the LCR calculated with these two operators are constantly better than LCR calculated with the two other operators in distinguishing wrong predictions from correct ones. Regarding the InceptionV3 CNN trained on the 100-classes ImageNet, we use both operators to generate mutants with the percentage of weight to change (mutation rate) arbitrarily set to 0.15\%, 0.2\% and 0.25\%. Wang \etal have not specified a recurrent method to apply in order to find the best mutation rate for any new model or new dataset but explain that as the mutation rate increases, the value difference between wrong predictions’ LCR and correct predictions’ LCR also increases, making the distinction clearer. However as we increase the percentage of weight to perturb, we create more unstable mutants that fail to achieve at least 90\% accuracy on the validation set, hence making the mutant generation phase longer. So we trade off by choosing the previous mutation rates as these values enable us to generate mutants in a reasonable amount of time and allow a good distinction between wrong and correct predictions. After generating 100 mutants per method and mutation rate, we calculate the LCR of samples from the 100-classes ImageNet dataset and display the average LCR and standard deviation of those samples in Table \ref{table_LCR}. As we want to optimize the LCR difference between correct and wrong predictions to be the largest possible, and NAI operator with a 0.2\% mutation rate having the largest LCR difference, we work only with this setting for the InceptionV3 CNN trained on the 100-classes ImageNet dataset. Regarding the number of mutants, intuitively the more mutants there are, the more accurate the LCR metric will be in distinguishing wrong and correct predictions. We limit the number of generated mutants to 100 because we observe that the capacity of the LCR metric to separate both classes as measured by the AUROC is capped around this number (see Figure \ref{figure_auroc_mn}). For the InceptionV3 CNN trained on the 50-classes ImageNet and the ResNet20 CNN trained on CIFAR10, we proceed in the same manner to finally decide using  NAI operator with a 0.2\% mutation rate and  NAI operator with a 0.1\% mutation rate respectively to generate 100 mutants for each dataset.

Using the generated mutants, we calculate LCR of images from training, validation and test sets according to equation \ref{equation_LCR} (step 8). The LCR of images vary from 0 to 1 so we label the LCR data as 0s or 1s for correct and wrong predictions. Since LCR are discriminant in themselves to classify predictions as wrong or correct, they do not need further processing. We use LCR of the three validation sets to display their respective ROC curves in Figure \ref{figure_allROCs}. From their ROC curves, we select thresholds using the same criteria of the conductance method (to have close performance on both classes of the validation set). We use these thresholds to classify LCR computed from the respective test set (step 10) and have the classification results displayed in Tables \ref{table_cifar10}, \ref{table_50imagenet} and \ref{table_100imagenet}.

\subsection{LCR and conductance as joint features}

\begin{figure*}[t]
\begin{center}
    \includegraphics[width=0.33\linewidth]{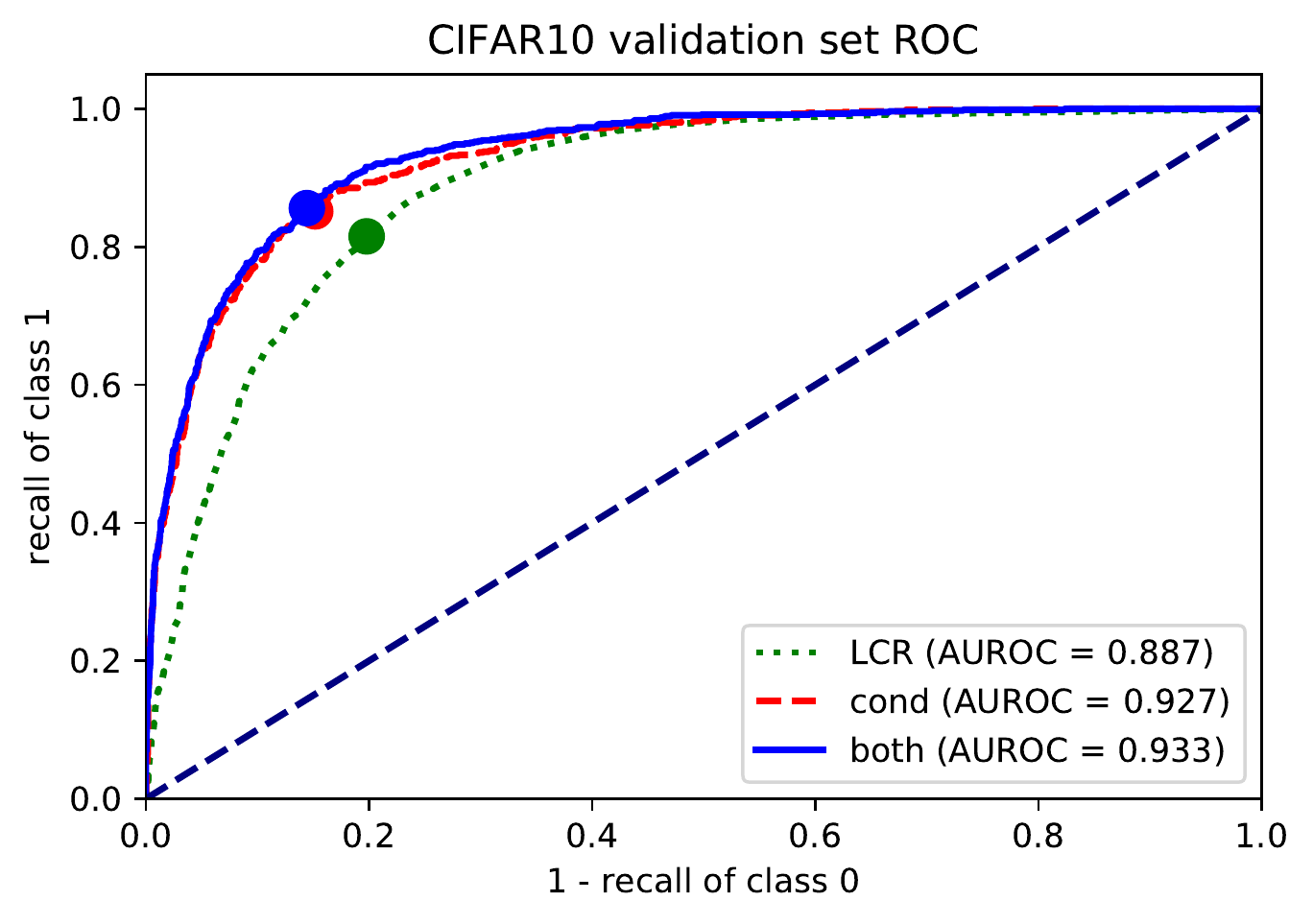}
    \includegraphics[width=0.33\linewidth]{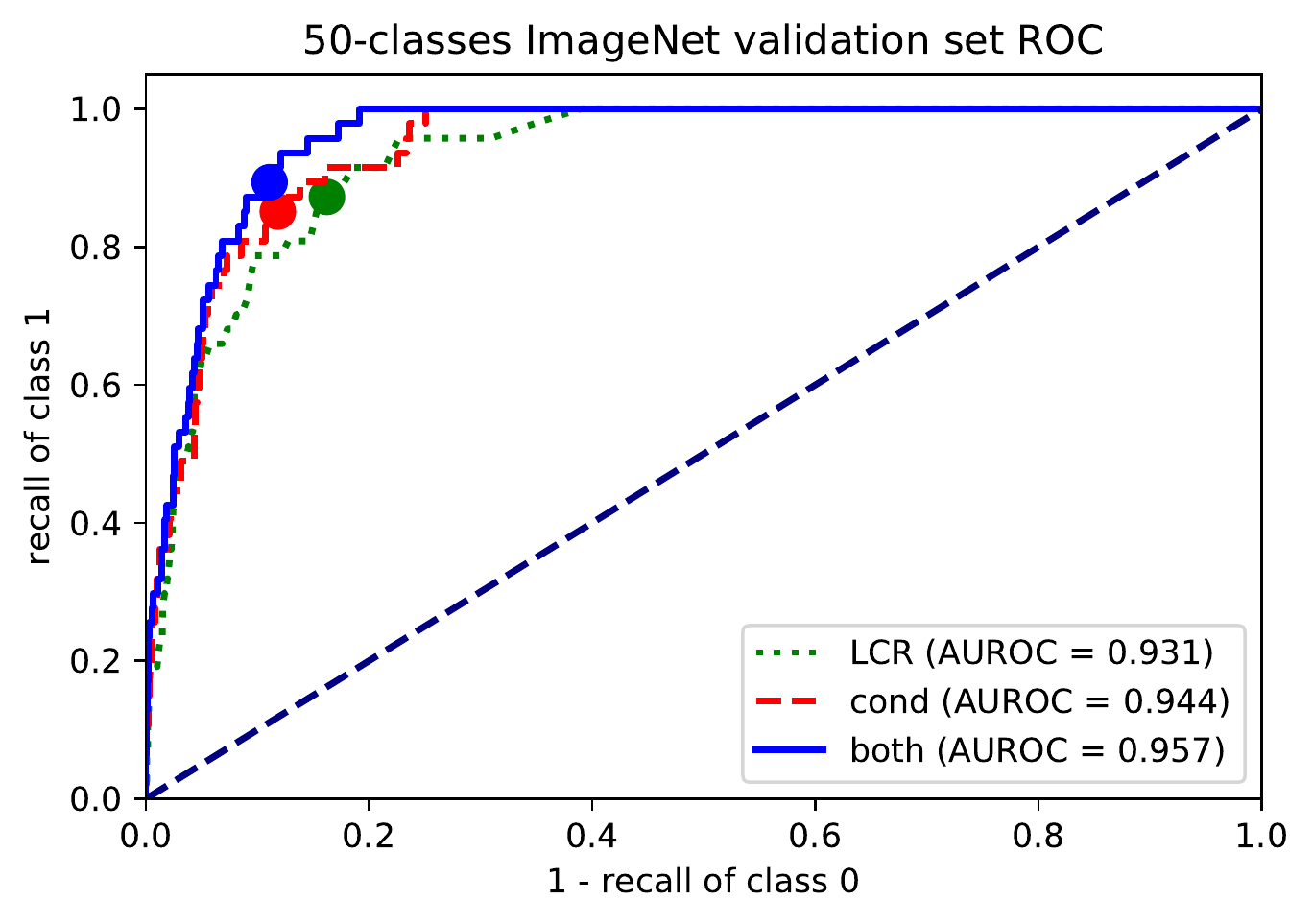}
    \includegraphics[width=0.33\linewidth]{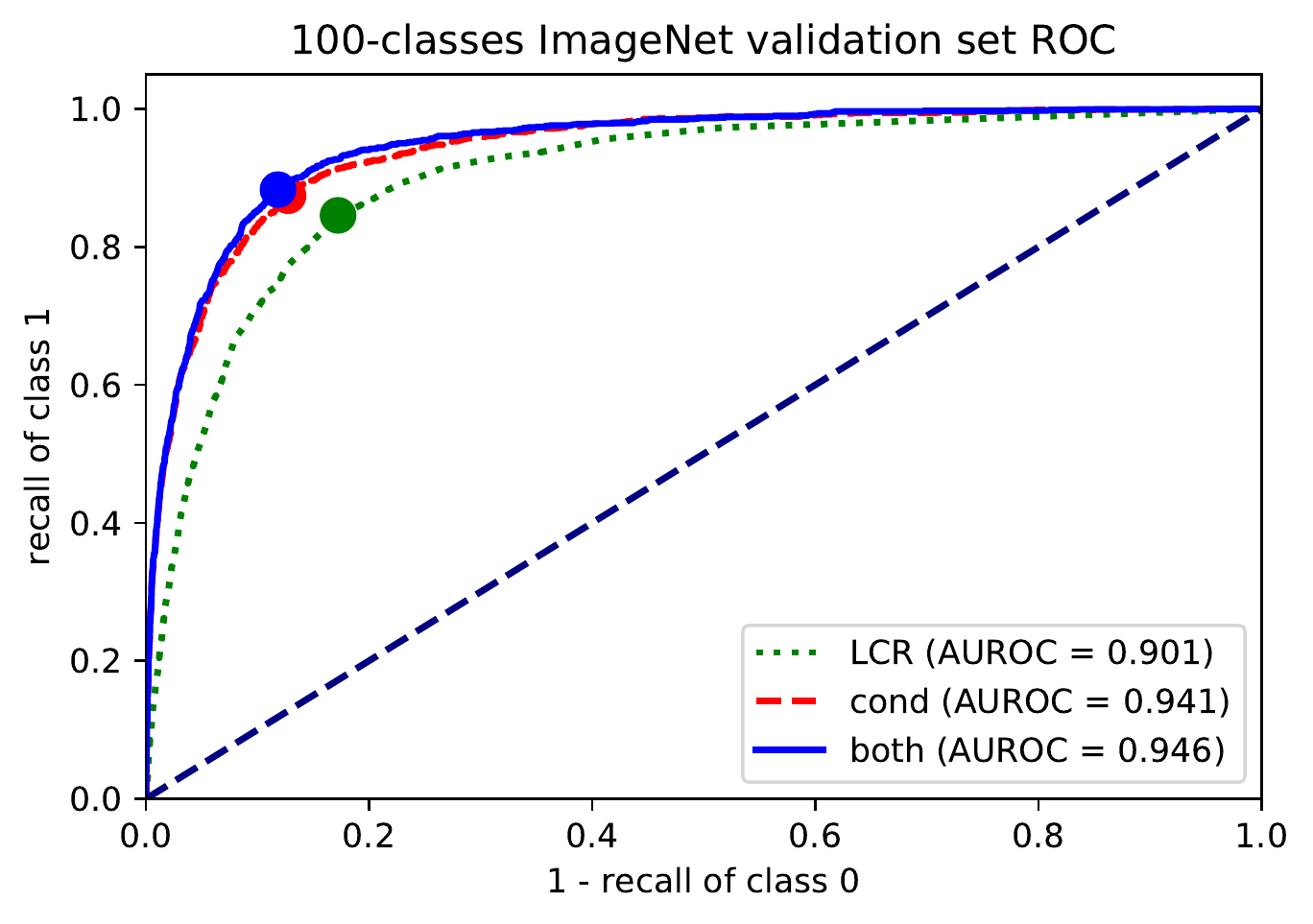}
\end{center}
   \caption{ROC curves measuring the performance of conductance, LCR and their combination in discriminating wrong and correct predictions of the three validation sets.}
\label{figure_allROCs}
\end{figure*}

While comparing the performances of LCR and conductance, we want to test if the combination of both of them could yield better results for discriminating wrong and correct predictions. On the one hand, the LCR of an input sample is a characteristic obtained by measuring the stability of its prediction w.r.t. the CNN parameters changes. On the other hand, the conductance of this sample is a characteristic obtained by tracking the activations pattern inside the CNN when processing this input. Since both metrics measure inherently different characteristics of the input sample, we suppose that considering both characteristics while taking a decision on the correctness of the input sample prediction would be more accurate than considering them separately. Experimental results verify our hypothesis as their combination constantly yields better performance (as measured by the AUROC) on the three datasets.

In order to combine LCR with conductance, we feed the output score of the conductance binary classifier and the LCR score to yet another unified binary classifier (step 11 of Figure \ref{figure_system}). The training, validation and test sets of the unified classifier is made up of concatenated LCR and conductance values computed from the image training, validation and test sets respectively. We observe an increase of the AUROC score for all validation sets and generally better performance on test sets (see Tables \ref{table_cifar10}, \ref{table_50imagenet} and \ref{table_100imagenet}). For the unified classifier we test solution like Neural Network (two hidden layers of 10 units each), Random Forest, Linear Discriminant Analysis and Quadratic Discriminant Analysis. We display only ROC curves obtained with the Neural Network model in Figure \ref{figure_allROCs}
as it performs as much if not better than the other models in all cases.

\begin{figure}[t]
\begin{center}
\includegraphics[width=0.6\linewidth]{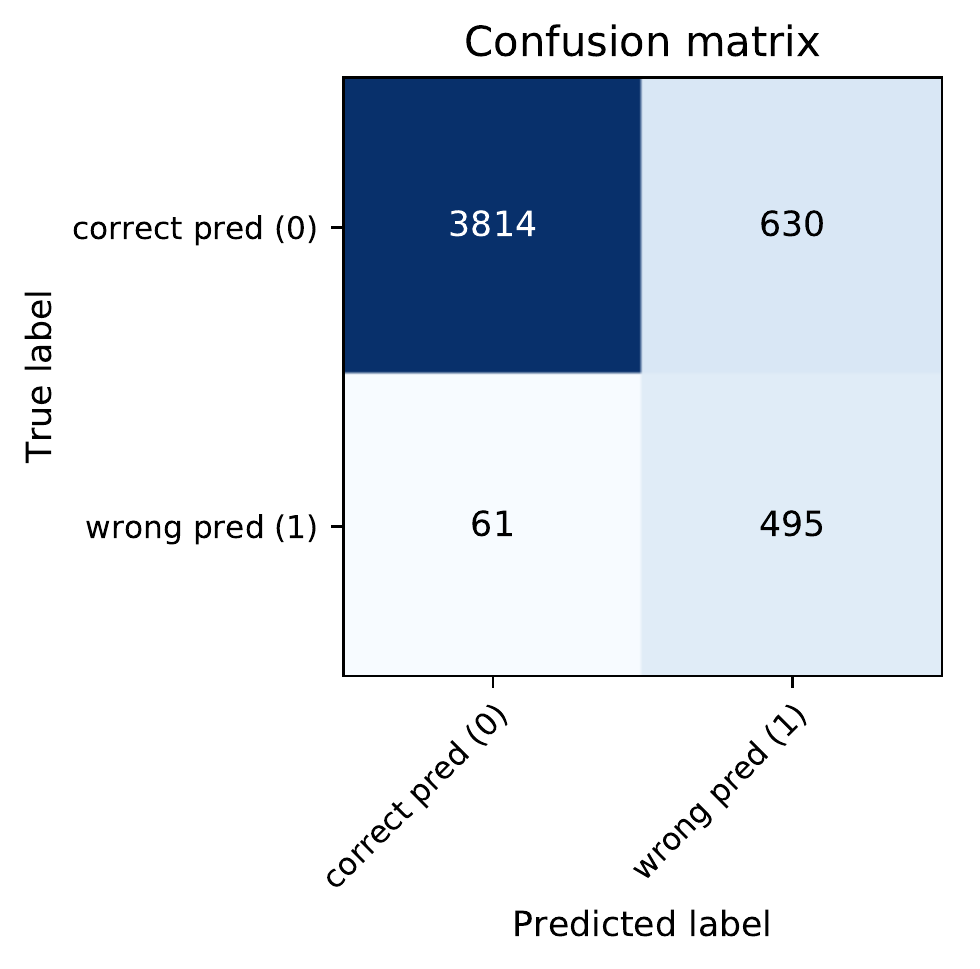}
\end{center}
\caption{Detailed classification results of the unified classifier on predictions of the 100-classes ImageNet test set.}
\label{figure_test_cm}
\end{figure}

In the case of the 100-classes ImageNet dataset, detailed results of the unified classifier's (Neural Network model) decisions to classify the test set predictions of the original CNN as wrong or correct (step 12 of Figure \ref{figure_system}) are displayed in Figure \ref{figure_test_cm}. With 5000 images in the test set, the original InceptionV3 CNN made 556 errors and 495 (89.0\% recall) of these errors are intercepted by the unified classifier. On the other hand, among 4444 correct predictions, it raised 630 false alarms by classifying part of the correct predictions as wrong (85.6\% recall). 

Previous results are obtained with a threshold that does not privilege any of the two classes. In another use case we might want the unified classifier to perform better on the class of correct prediction (0) than the class of wrong prediction (1). For the 100-classes ImageNet dataset, we choose another threshold to have 98.2\% and 50.4\% recall for correct and wrong prediction classes on the validation set. We use this threshold to classify predictions made by the CNN on the test set. We obtain 97.3\% and 55.9\% recall for correct and wrong prediction classes. Among 556 errors made by the original CNN, it is then 310 wrong predictions that are intercepted by the unified classifier and among 4444 correct predictions of the CNN, the unified classifier then considers 120 of them as wrong predictions.

In the opposite case, we want to minimize residual wrong predictions. We can choose a threshold resulting in 43.0\% and 99.5\% recall for correct and wrong prediction classes on the test set.

\begin{table}[t]
\centering
\scalebox{0.60}{
\begin{tabular}{|c|c|c|}
\hline
& validation set& test set\\
\hline
\begin{tabular}{c}
     Setting \\
     \hline
     Conductance\\
     LCR \\
     \hline
     NN both \\
     RF both\\
     LDA both\\
     QDA both\\
\end{tabular}&
\begin{tabular}{c|c|c}
     AUROC& correct recall&wrong recall  \\
     \hline
     0.927&0.851&0.851 \\
     0.887&0.815&0.802\\
     \hline
     \textbf{0.933}&\textbf{0.856}&\textbf{0.856}\\
     0.930&0.853&0.854\\
     0.932&\textbf{0.856}&\textbf{0.856}\\
     0.928&0.854&0.854\\
\end{tabular}&
\begin{tabular}{c|c}
     correct recall& wrong recall  \\
     \hline
     0.863&0.850  \\
     0.841&0.806 \\
     \hline
     \textbf{0.870}&\textbf{0.861}\\
     0.855&0.858\\
     0.862&0.859\\
     0.844&\textbf{0.861}\\
\end{tabular}\\
\hline
\end{tabular}}
\caption{CIFAR10, results of different settings to classify predictions made by the ResNet20 with a threshold to enable close recall scores on the validation set. NN: Neural Network, RF: Random Forest, LDA: Linear Discriminant Analysis, QDA: Quadratic Discriminant Analysis.}
\label{table_cifar10}
\end{table}
\begin{table}[t]
\centering
\scalebox{0.60}{
\begin{tabular}{|c|c|c|}
\hline
& validation set& test set\\
\hline
\begin{tabular}{c}
     Setting \\
     \hline
     Conductance\\
     LCR \\
     \hline
     NN both \\
     RF both\\
     LDA both\\
     QDA both\\
\end{tabular}&
\begin{tabular}{c|c|c}
     AUROC& correct recall&wrong recall  \\
     \hline
     0.944&0.873&0.872 \\
     0.931&0.844&0.872\\
     \hline
     \textbf{0.957}&0.891&\textbf{0.894}\\
     0.948&0.860&0.872\\
     0.954&0.877&0.872\\
     0.954&\textbf{0.893}&\textbf{0.894}\\
\end{tabular}&
\begin{tabular}{c|c}
     correct recall& wrong recall  \\
     \hline
     0.820&0.886  \\
     0.816&0.943 \\
     \hline
     0.842&\textbf{0.960}\\
     0.810&0.943\\
     0.832&0.949\\
     \textbf{0.852}&0.937\\
\end{tabular}\\
\hline
\end{tabular}}
\caption{50-Classes ImageNet, results of different settings to classify predictions made by the InceptionV3 with a threshold to enable close recall scores on the validation set.}
\label{table_50imagenet}
\end{table}
\begin{table}[t]
\centering
\scalebox{0.60}{
\begin{tabular}{|c|c|c|}
\hline
& validation set& test set\\
\hline
\begin{tabular}{c}
     Setting \\
     \hline
     Conductance\\
     LCR \\
     \hline
     NN both \\
     RF both\\
     LDA both\\
     QDA both\\
\end{tabular}&
\begin{tabular}{c|c|c}
     AUROC& correct recall&wrong recall  \\
     \hline
     0.941&0.873&0.873 \\
     0.901&0.836&0.831\\
     \hline
     \textbf{0.946}&\textbf{0.882}&\textbf{0.881}\\
     0.943&0.875&0.875\\
     0.944&0.875&0.875\\
     0.942&0.874&0.873\\
\end{tabular}&
\begin{tabular}{c|c}
     correct recall& wrong recall  \\
     \hline
     0.847&\textbf{0.896}  \\
     0.812&0.856 \\
     \hline
     \textbf{0.858}&0.890\\
     0.847&0.903\\
     0.849&0.894\\
     0.852&0.890\\
\end{tabular}\\
\hline
\end{tabular}}
\caption{100-classes ImageNet, results of different settings to classify predictions made by the InceptionV3 with a threshold to enable close recall scores on the validation set.}
\label{table_100imagenet}
\end{table}

\section{Discussion}

\textbf{Our method's utilities}, predictions classified as wrong are not further processed in our case. It creates one more class of prediction which is the unknown class for discarded samples and the CNN alone cannot take decisions on predictions of this class. However the method can still be useful in multi-modal systems. Input samples on which the CNN did not take decision can be relayed to other modules for them to decide what to do with these samples. In autonomous car, we can think of LIDAR that can detect an object in front of the car if the CNN is not sure if it detected an object. For operations that need human supervision such as medical diagnosis or malware detection, inputs classified as unknown by a CNN can be processed directly by specialists creating an interaction between humans and machines. In this case, let us suppose we have the results obtained on the 100-classes ImageNet which is an original CNN with a 88.9\% accuracy that classified 4444 images correctly among 5000. The unified classifier would discard 630 inputs among 4444 correctly predicted inputs and 495 inputs among 556 wrongly predicted inputs making a total of 1125 discarded inputs. If we suppose that the specialists could rectify all of the 1125 inputs correctly, the overall prediction accuracy on the 5000 inputs becomes 98.4\%. 

Another option is when we want to make rapid inference in a system with a small CNN and have a more accurate but slower and bigger CNN as backup. The system takes most of the decisions with the small CNN and when there are discarded inputs that are classified as unknown by the small CNN, it relays these inputs to the bigger CNN for it to make decisions. 

Our method can also help DNNs address safety concepts for their deployment in DNN-based software architectures. The safety concepts are such that the overall system needs to be able to handle or mitigate unexpected behaviors of the DNN. To do so, additional software components working in parallel to the DNN can be used to check the model output for consistency, detect abnormal neural behaviors and check whether the input is statistically close to the training dataset. By nature, our method hence can be used as a solution to satisfy AI safety purposes.

\textbf{Metrics comparison}, generally we observe that conductance performs better than LCR. Wang \etal state that the more accurate the original CNN is, the more accurate the LCR will be in distinguishing wrong and correct predictions. While for conductance, we observe that its performance is relatively stable across CNNs and datasets. To explain the higher performance gain for the 50-classes ImageNet when combining both methods, because the CNN trained on this dataset had better accuracy than CNNs trained on the two other datasets, LCR performs better on the 50-classes ImageNet CNN, returning a bigger contribution when combined with conductance (see Figure~\ref{figure_allROCs}).

\textbf{Inference computational cost}, to obtain the LCR information of an input, we need to make 100 inferences on same sized CNNs (100 mutants making predictions, see equation \ref{equation_LCR}) while the obtention of conductance needs only 11 inferences (as the Riemann approximation step $n$ is 10, see equation \ref{equation_fastcond_approx}) plus additional back propagation operations. Subsequently, the conductance method is faster than LCR for deciding if the prediction of a CNN is wrong or correct.

\section{Conclusion}

Nowadays there is an abundant number of researches around the subject of Deep Neural Networks (DNN) interpretability. Since DNNs are known to have complex underlying inference mechanism, these researches aim to provide a better understanding of DNNs in order to be able to bring corrections to them. However research around the subject of improving DNN performance with interpretability methods is still limited and we contribute to this area of research in this paper. We show that an interpretability method can be used to reduce errors made by a CNN hence increasing its robustness. To increase the CNN's robustness, while our approach consider it as a white box by using an interpretability method, a black box approach of the problematic which does not use interpretability methods also exists in the literature. We compare both methods and show that since they rely on different resources, using them jointly can achieve better results.  

{\small
\bibliographystyle{ieee}
\bibliography{egbib}
}

\end{document}